\documentclass[runningheads]{llncs}
\usepackage[T1]{fontenc}
\usepackage{graphicx}
\usepackage{epsfig}
\usepackage{graphicx}
\usepackage{amsmath}
\usepackage{amssymb}

\usepackage{booktabs}
\usepackage{bm}
\usepackage{amssymb}
\usepackage{pifont}
\usepackage{caption}
\usepackage{multirow} 
\usepackage{xcolor,colortbl}
\usepackage{tabularx,booktabs}

\usepackage{hyperref}
\hypersetup{
    colorlinks=true,
    linkcolor=blue,
    filecolor=magenta,      
    urlcolor=blue,
    pdftitle={SSL4EVA - Regev Cohen},
    pdfpagemode=FullScreen,
    }

\newcommand{\changes}[1]{\textcolor{black}{#1}}

\definecolor{ffe1da}{RGB}{255,225,218}
\definecolor{F7E0D5}{RGB}{247,224,213}
\definecolor{mycolor}{RGB}{173,216,230}
\definecolor{darkF7E0D5}{RGB}{209,154,128}
\definecolor{White}{RGB}{255,255,255}
\colorlet{Light}{White!80!mycolor}

\def\vp{{\mathbf{p}}}

\def\vx{{\mathbf{x}}}

\def\vz{{\mathbf{z}}}

\def\mQ{{\mathbf{Q}}}

\def\sR{{\mathbb{R}}}

\begin{document}
\title{Self-Supervised Learning \\ for Endoscopic Video Analysis}
\titlerunning{Self-Supervised Learning for Endoscopy}
%
\author{Roy Hirsch\inst{1} \and Mathilde Caron\inst{2} \and Regev Cohen\inst{1}\thanks{Corresponding Author: \email{regevcohen@google.com}} 
\and Amir Livne\inst{1} \and Ron Shapiro\inst{1} \and Tomer Golany\inst{1} \and Roman Goldenberg\inst{1} \and Daniel Freedman\inst{1} \and Ehud Rivlin\inst{1}
}

%
\authorrunning{R. Hirsch et al.}
%
\institute{Verily AI \and Google Research
}

\maketitle              
\begin{abstract}
Self-supervised learning (SSL) has led to important breakthroughs in computer vision by allowing learning from large amounts of \textit{unlabeled} data.
As such, it might have a pivotal role to play in biomedicine where annotating data requires a highly specialized expertise.
Yet, there are many healthcare domains for which SSL has not been extensively explored. One such domain is endoscopy, minimally invasive procedures which are commonly used to detect and treat infections, chronic inflammatory diseases or cancer. In this work, we study the use of a leading SSL framework, namely Masked Siamese Networks (MSNs), for endoscopic video analysis such as colonoscopy and laparoscopy. To fully exploit the power of SSL, we create sizable \changes{\textit{unlabeled} endoscopic video datasets for training MSNs. These strong image representations serve as a foundation for secondary training with limited annotated datasets, resulting in state-of-the-art performance in endoscopic benchmarks like surgical phase recognition during laparoscopy and colonoscopic polyp characterization. Additionally, we achieve a 50\% reduction in annotated data size without sacrificing performance.} Thus, our work provides evidence that SSL can dramatically reduce the need of annotated data in endoscopy. 

\keywords{Artificial intelligence  \and Self-Supervised Learning \and Endoscopy Video Analysis.}

\end{abstract}
\section{Introduction}
\label{sec:intro}
Endoscopic operations are minimally invasive medical procedures which allow physicians to examine inner body organs and cavities. During an endoscopy, a thin, flexible tube with a tiny camera is inserted into the body through a small orifice or incision. It is used to diagnose and treat a variety of conditions, including ulcers, polyps, tumors, and inflammation.
Over 250 million endoscopic procedures are performed each year globally and 80 million in the United States,
signifying the crucial role of endoscopy in clinical research and care.

A cardinal challenge in performing endoscopy is the limited field of view which hinders navigation and proper visual assessment, potentially leading to high detection miss-rate, incorrect diagnosis or insufficient treatment. These limitations have fostered the development of computer-aided systems based on artificial intelligence (AI), 
resulting in unprecedented performance over a broad range of clinical applications \cite{golany2022artificial,livovsky2021detection,kutiel2023conformal,katzir2022estimating,cohen2021has,cohen2021regularization}. Yet the success of such AI systems heavily relies on acquiring annotated data which requires experts of specific knowledge, leading to an expensive, prolonged process. 
In the last few years, Self-Supervised Learning (SSL \cite{chen2021exploring,caron2020unsupervised,caron2021emerging,chen2020simple}) has been shown to be a revolutionary strategy for unsupervised representation learning, eliminating the need to manually annotate vast quantities of data. Training large models on sizable unlabeled data via SSL leads to powerful representations which are effective for downstream tasks with few labels. However, research in endoscopic video analysis has only scratched the surface of SSL which remains largely unexplored. 

This study introduces Masked Siamese Networks (MSNs \cite{assran2022masked}), a prominent SSL framework, into endoscopic video analysis where we focus on laparoscopy and colonoscopy. We first experiment solely on public datasets, Cholec80 \cite{twinanda2016endonet} and PolypsSet \cite{polypset2021},
demonstrating performance on-par with the top results reported in the literature. Yet, the power of SSL lies in large data regimes. Therefore, to exploit MSNs to their full extent, we collect and build two sizable \changes{\textit{unlabeled}} datasets for laparoscopy and colonoscopy with $7,700$ videos (>23M frames) and $14,000$ videos (>2M frames) respectively. Through extensive experiments, we find that scaling the data size necessitates scaling the model architecture, leading to state-of-the-art performance in surgical phase recognition of laparoscopic procedures, as well as in polyp characterization of colonoscopic videos. Furthermore, the proposed approach exhibits robust generalization, yielding better performance with only 50\% of the annotated data, \changes{compared with standard supervised learning using the complete labeled dataset.} This shows the potential to reduce significantly the need for expensive annotated medical data. 


\section{Background and Related Work}
\label{sec:background}
There exist a wide variety of endoscopic applications. Here, we focus on colonoscopy and laparoscopy, which combined covers over 70\% of all endoscopic procedures. Specifically, our study addresses two important common tasks, described below. 

\paragraph{\textbf{Cholecystectomy Phase Recognition}} 
Cholecystectomy is the surgical removal of the gallbladder using small incisions and specialized instruments. 
It is a common procedure performed to treat gallstones, inflammation, or other conditions affecting the gallbladder. Phase recognition in surgical videos is an important task that aims to improve surgical workflow and efficiency. Apart from measuring quality and monitoring adverse event, this task also serves in facilitating education, statistical analysis, and evaluating surgical performance. Furthermore, the ability to recognize phases allows real-time monitoring and decision-making assistance during surgery, thus improving patient safety and outcomes. AI solutions have shown remarkable performance in recognizing surgical phases of cholecystectomy procedures~\cite{goldbraikh2023bounded,golany2022artificial,twinanda2016endonet};
however, they typically require large labelled training datasets. As an alternative, SSL methods have been developed~\cite{ross2018exploiting,da2019self,sestini2021kinematic},
however, these are early-days methods that based on heuristic, often require external information and leads to sub-optimal performance. A recent work~\cite{ramesh2023dissecting} presented an extensive analysis of modern SSL techniques for surgical computer vision, yet on relatively small laparoscopic datasets.

\paragraph{\textbf{Optical Polyp Characterization}} 
\label{subsec:colo}
Colorectal cancer (CRC) remains a critical health concern and significant financial burden worldwide.
Optical colonoscopy
is the standard of care screening procedure for preventing CRC through the identification and removal of polyps~\cite{byrne2017will}. According to colonoscopy guidelines, all identified polyps must be removed and histologically evaluated regardless of their malignant nature. 
Optical biopsy enables practitioners to remove pre-cancerous adenoma polyps or leave distal hyperplastic polyps in situ without the need for pathology examination, by visually predicting histology. However, this technique is highly dependent on operator expertise \cite{dayyeh2015asge}.
This limitation has motivated the development of AI systems for automatic optical biopsy, allowing  non-experts to also effectively perform optical biopsy during polyp management. In recent years, various AI systems have been developed to this end \cite{antonelli2023current,hassan2021performance}.
However, training such automatic optical biopsy systems relies on a large body of annotated data, while SSL has not been investigated in this context, to the best of our knowledge.  


\section{Self-Supervised Learning for Endoscopy}
\label{sec:method}
\begin{figure*}[htbp]
  \centering
  \includegraphics[width=\textwidth]{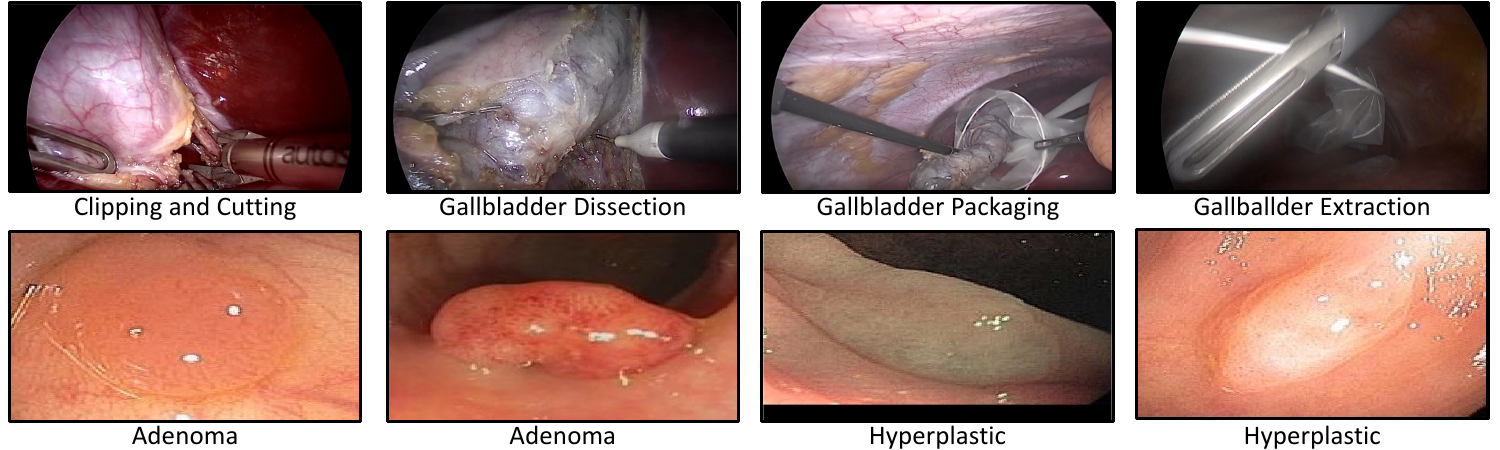}
  \caption{\textbf{Data Samples.} Top: Laparoscopy. Bottom: Colonoscopy.}
  \label{fig:data-samples}
\end{figure*}

SSL approaches have produced impressive results recently~\cite{chen2020simple,caron2021emerging,chen2021exploring,caron2020unsupervised}, relying on two key factors: (i) effective algorithms for unsupervised learning and (ii) training on large-scale datasets.
Here, we first describe Masked Siamese Networks~\cite{assran2022masked}, our chosen SSL framework. Additionally, we present our large-scale data collection (see Fig.~\ref{fig:data-samples}). Through extensive experiments in Sec.~\ref{sec:exp}, we show that training MSNs on these substantial datasets unlocks their potential, yielding effective representations that transfer well to public laparoscopy and colonoscopy datasets.



\subsection{Masked Siamese Networks}
\label{subsec:msn}
SSL has become an active research area, giving rise to efficient learning methods such as SimCLR~\cite{chen2020simple}, SwAV~\cite{caron2020unsupervised} and DINO~\cite{caron2021emerging}.
Recently, Masked Siamese Networks~\cite{assran2022masked} have set a new state-of-the-art among SSL methods on the ImageNet benchmark~\cite{russakovsky2015imagenet}, with a particular focus on the low data regime.
This is of great interest for us since our downstream datasets are typically of small size~\cite{twinanda2016endonet,polypset2021}. 
We briefly describe MSNs below and refer the reader to \cite{assran2022masked} for further details.

\begin{figure*}[htbp]
  \centering
  \includegraphics[trim={2cm 4cm 2cm 5cm}, clip, height=3.5cm, width=\textwidth]{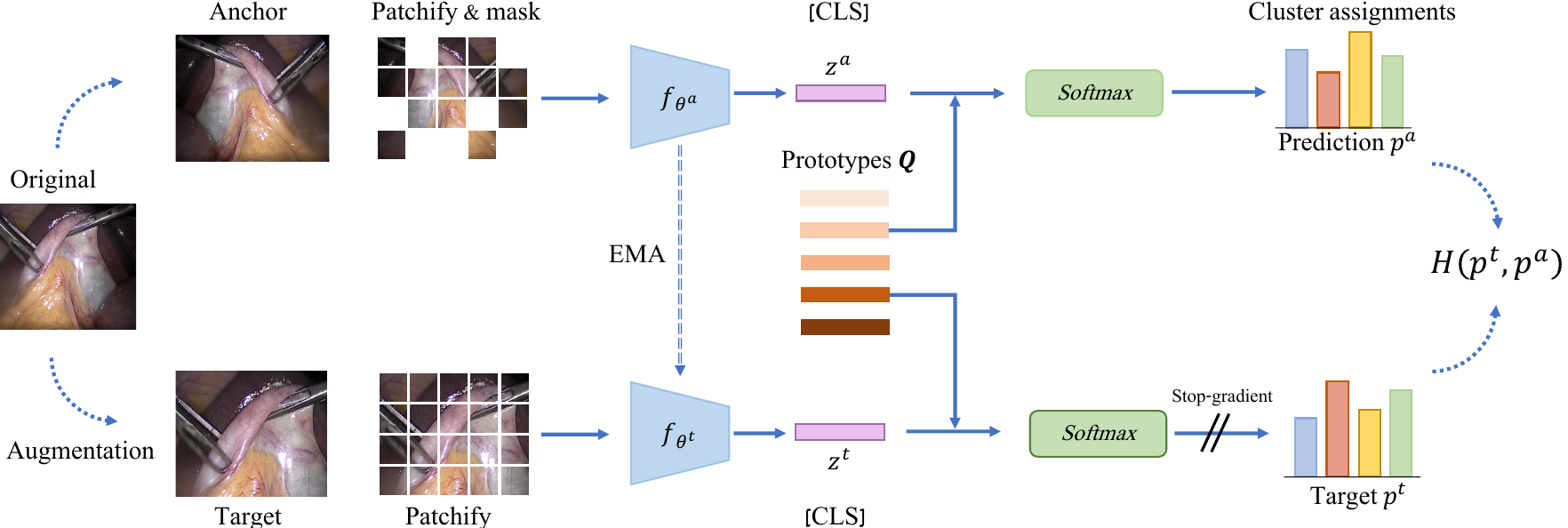}
  \caption{\textbf{Schematic of Masked Siamese Networks.}}
  \label{fig:msn}
\end{figure*}

During pretraining, on each image $\vx_i\in\sR^n$ of a mini-batch of $B\geq 1$ samples (e.g. laparoscopic images) we apply two sets of random augmentations to generate anchor and target views, denoted by $\vx_i^a$ and $\vx_i^t$ respectively.
We convert each view into a sequence of non-overlapping patches and perform an additional masking (``random'' or ``focal'' styles) step on the anchor view by randomly discarding some of its patches. The resultant anchor and target sequences are used as inputs to their respective image encoders $f_{\theta^a}$ and $f_{\theta^t}$. Both encoders share the same Vision Transformer (ViT~\cite{dosovitskiy2020image}) architecture where the parameters $\theta^t$ of the target encoder
are updated via an exponential moving average of the anchor encoder parameters $\theta^a$. The outputs of the networks are the representation vectors $\vz_i^a\in\sR^d$ and $\vz_i^t\in\sR^d$, corresponding to the [CLS] tokens of the networks. 
The similarity between each view and a series of $K>1$ learnable prototypes is then computed, and the results undergo a softmax operation to yield the following probabilities $\vp_i^a= softmax\left(\frac{\mQ\vz_i^a}{\tau^a}\right)$ and $\vp_i^t= softmax\left(\frac{\mQ\vz_i^t}{\tau^t}\right)$
where $0<\tau^t<\tau^a<1$ are temperatures and $\mQ\in\sR^{K\times d}$ is a matrix whose rows are the prototypes. The probabilities are promoted to be the same by minimizing the cross-entropy loss $H(p_i^t,p_i^a)$, as illustrated in Fig.~\ref{fig:msn}. 

In practice, a sequence of $M\geq1$ anchor views are generated, leading to multiple probabilities $\{\vp_{i,m}^a\}_{m=1}^M$. Furthermore, to prevent representation collapse and encourage the model to fully exploit the prototypes, a mean entropy maximization (me-max) regularizer \cite{assran2022masked,joulin2012convex} is added, aiming to maximize the entropy $H(\bar\vp^a)$ of the average prediction across all the anchor views $\bar\vp^a\triangleq \frac{1}{MB}\sum_{i=1}^B\sum_{m=1}^M\vp^a_{i,m}$.
Thus, the  overall training objective to be minimized for both $\theta^a$ and $\mQ$ is
where $\lambda>0$ is an hyperparameter and the gradients are computed only with respect to the anchor predictions $\vp^a_{i,m}$ (not the target predictions $\vp_i^t$).
Applying MSNs on the large datasets described below, generates representations that serve as a strong basis for various downstream tasks, as shown in the next section.

\subsection{Private Datasets}
\label{subsec:data}

\paragraph{\textbf{Laparoscopy.}} 
\changes{
We compiled a dataset of laparoscopic procedures videos exclusively performed on patients aged 18 years or older.  The dataset consists of 7,877 videos recorded at eight different medical centers in Israel. The dataset predominantly consists of the following procedures: cholecystectomy (35\%), appendectomy (20\%), herniorrhaphy (12\%), colectomy (6\%), and bariatric surgery (5\%). The remaining 21\% of the dataset encompasses various standard laparoscopic operations.
The recorded procedures have an average duration of 47 minutes, with a median duration of 40 minutes. Each video recording was sampled at a rate of 1 frame per second (FPS), resulting in an extensive dataset containing 23.3 million images. Further details are given in the supplementary materials.}  


\paragraph{\textbf{Colonoscopy.}} 
\changes{
We have curated a dataset comprising 13,979 colonoscopy videos of patients aged 18 years or older. These videos were recorded during standard colonoscopy procedures performed at six different medical centers between the years 2019 and 2022. The average duration of the recorded procedures is 15 minutes, with a median duration of 13 minutes.
To identify and extract polyps from the videos, we employed a pretrained polyp detection model \cite{livovsky2021detection,ou2021polyp,intrator2023reid}. Using this model, we obtained bounding boxes around the detected polyps. To ensure high-quality data, we filtered out detections with confidence scores below 0.5. For each frame, we cropped the bounding boxes to generate individual images of the polyps. This process resulted in a comprehensive collection of 2.2 million polyp images.}


\section{Experiments} \label{sec:exp}
In this section, we empirically demonstrate the power of SSL in the context of endoscopy.
Our experimental protocol is the following:
(i)~first, we perform \textit{SSL pretraining} with MSNs over our \changes{unlabeled} private dataset to learn informative and generic representations,
(ii)~second we probe these representations by utilizing them for different public \textit{downstream tasks}.
Specifically, we use the following two benchmarks.
(a) \textit{Cholec80}~\cite{twinanda2016endonet}: 80 videos of  cholecystectomy procedures resulting in nearly 200k frames at 1 FPS. Senior surgeons annotated each frame to one out of seven phases.
(b) \textit{PolypsSet}~\cite{polypset2021}: A unified dataset of 155 colonoscopy videos (37,899 frames) with labeled polyp classes (hyperplastic or adenoma) and bounding boxes. We use the provided detections to perform binary classification.

\paragraph{\textbf{Downstream Task Evaluation Protocols.}}
(a) {\textit{Linear evaluation:}} A standard protocol consisting in learning a linear classifier on top of frozen SSL features~\cite{caron2021emerging,he2022masked}. (b) \textit{Temporal evaluation:} A natural extension of the linear protocol where we learn a temporal model on top of the frame-level frozen features.
We specifically use Multi-Stage Temporal Convolution Networks (MS-TCN) as used in~\cite{czempiel2020tecno,ramesh2023dissecting}.
\changes{This incorporates the temporal context which is crucial for video tasks such as phases recognition.} (c) \textit{Fine-tuning:} An end-to-end training of a classification head on top of the (unfrozen) pretrained backbone. We perform an extensive hyperparameter grid search for all downstream experiments and report the test results for the models that exceed the best validation results. 
We report the Macro F1 (F-F1) as our primary metric. For phase recognition we also report the per-video F1 (V-F1), computed by averaging the F1 scores across all videos~\cite{ramesh2023dissecting}.

\paragraph{\textbf{Implementation Details.}}
For SSL we re-implemented MSNs in JAX using Scenic library~\cite{dehghani2021scenic}. As our image encoders we train Vision Transformer (ViT~\cite{dosovitskiy2020image}) of different sizes, abbreviated as ViT-S/B/L, using 16 TPUs. Downstream experiments are implemented in TensorFlow
where training is performed on 4 Nvidia Tesla V100 GPUs. See the supplementary for further implementation details.\footnote{\changes{For reproducibility purposes, code and model checkpoints are available at} \url{https://github.com/RoyHirsch/endossl}}


\subsection{Results and Discussion}
\paragraph{\textbf{Scaling laws of SSL.}}
We explore large scale SSL pretraining for endoscopy videos.
Table~\ref{tab:scaling} compares the results of pretraining with different datasets (public and private) and model sizes. We pretrain the models with MSN and then report their downstream performances. We present results for the cholecystectomy phase recognition task based on fine-tuned models and for the optical polyp characterization task based on linear evaluation, due to the small size of the public dataset. \changes{As baselines, we report fully-supervised ResNet50 results, trained on public datasets. We find that replacing ResNet50 with ViT-S,  despite comparable  number of parameters, yields sub-optimal performance.}

 SSL pretraining on public datasets \changes{(without labels)} provides comparable or better results than fully supervised baselines. The performance in per-frame phase recognition is comparable with the baseline. Phase recognition per-video results improve by 1.3 points when using the MSN pretraining, while polyp characterization improve by 2.2 points.
Importantly, we see that the performance gap becomes prominent when using the large scale private datasets for SSL pretraining. Here, per-frame and per-video phase recognition performances improve by $6.7\%$ and $8.2\%$, respectively. When using the private colonoscopy dataset the Macro F1 improves by $11.5\%$ compared to the fully supervised baseline. 
Notice that the performance improves with scaling both model and private data sizes, demonstrating that both factors are crucial to achieve optimal performance. 

\begin{table}[]
    \scriptsize
    \centering
    \caption{Comparing the downstream F1 performances of: 
    (i) Models trained on the private (Pri) and public (Pub) datasets using SSL. (ii) Fully supervised baselines pretrained on ImageNet-1K (IN1K). Best results are highlighted.}
    \begin{tabular}{lcc|c|cc|c}
    \toprule
        \multirow{2}{*}{Method} & \multirow{2}{*}{Arch} & \multirow{2}{*}{Pretrain} & \multirow{2}{*}{Cholec80 frame} & \multicolumn{2}{c|}{Cholec80 temporal} & \multirow{2}{*}{PolypsSet} \\ 
        ~ & ~ & ~ & ~ & F-F1 & V-F1 & ~ \\ \midrule
        \multicolumn{7}{l}{\textit{Fully Supervised}} \\ \midrule
        FS~\cite{ramesh2023dissecting} & RN50 & IN1K & 71.5 & - & 80.3 & 72.1\\
        TeCNO & RN50 & IN1K & - & 83.3 & - & - \\ 
        OperA & RN50 & IN1K & - & 84.4 & - & - \\ \midrule
        \multicolumn{7}{l}{\textit{Self Supervised}} \\ \midrule
        DINO & ViT-S  & IN1K & 64.9 & 77.4 &72.4 & 61.0 \\
        DINO~\cite{ramesh2023dissecting} & RN50 & Pub & 71.1 & - & 81.6 & 72.4\\ 
        MSN & ViT-S & Pub & 65.0 & 83.4 & 80.9 & 70.6\\
        MSN & ViT-B & Pub & \colorbox{Light}{71.2} & 82.6 & \colorbox{Light}{82.9} & \colorbox{Light}{74.6}\\
        MSN & ViT-L & Pub & 65.6 & \colorbox{Light}{84.0} & 82.0 & 73.6\\  \midrule
        MSN & ViT-S & Pri & 70.7 & 87.0 & 84.3 & 78.5\\
        MSN & ViT-B & Pri & 73.5 & 87.3 & 85.8 & 78.2\\
        MSN & ViT-L & Pri & \colorbox{Light}{76.3} & \colorbox{Light}{89.6} & \colorbox{Light}{86.9} & \colorbox{Light}{80.4}\\ 
    \bottomrule
    \end{tabular}
\label{tab:scaling}
\end{table}

\begin{figure*}[t]
  \centering
  \includegraphics[height=3cm, width=\textwidth]{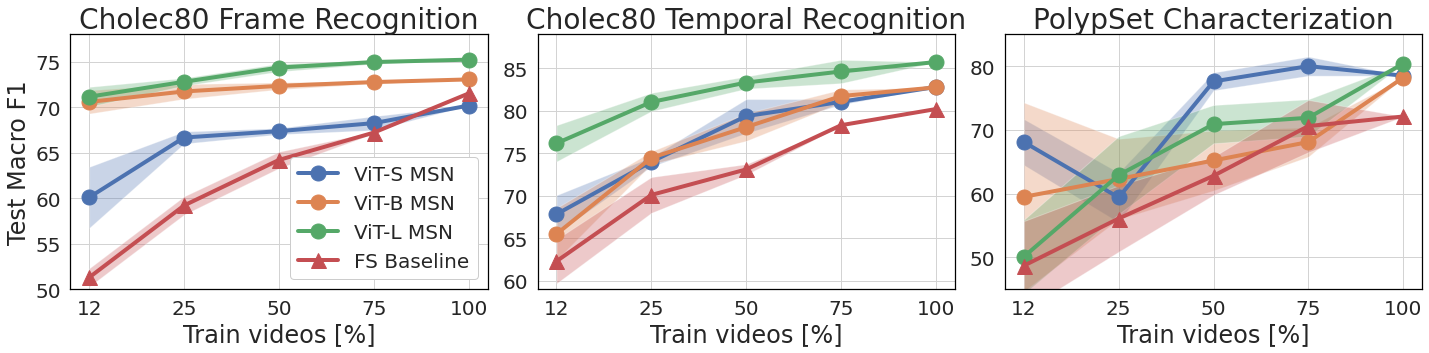}
  \caption{Low-shot evaluation comparing MSN to fully supervised baselines.}
  \label{fig:low_data}
\end{figure*}

\paragraph{\textbf{Low-Shot Regime.}}
Next, we examine the benefits of using MSNs to improve downstream performance in a \textit{low-shot} regime with few annotated samples.
Note that MSNs have originally been found to produce excellent features for low data regime~\cite{assran2022masked}.
We train a linear classifier on top of the extracted features and report the test classification results. Figure~\ref{fig:low_data} shows the low-shot performance for the two endoscopic tasks. We report results using a fraction $k=\{12\%, 25\%, 50\%, 75\%, 100\%\}$ of the annotated \changes{public} videos. We also report results for fully-supervised baselines trained on the same fraction of annotated samples. Each experiment is repeated three times with a random sample of train videos, and we report the mean and standard deviation (shaded area).

As seen, SSL-based models provide enhanced robustness to limited annotations. When examining the cholecystectomy phase recognition task, it is evident that we can achieve comparable frame-level performance by using only $12\%$ of the annotated videos. Using $25\%$ of the annotated videos yields comparable results to the fully supervised temporal models. Optical polyp characterization results show a similar trend, but with a greater degree of variability. \changes{Using small portions of PolypSet (12\% and 25\%) hindered the training process and increased sensitivity to the selected portions. However, when using more than 50\% of PolypSet, the training process stabilized, yielding results comparable to the fully supervised baseline. This feature is crucial for medical applications, given the time and cost involved in expert-led annotation processes.}
%

\subsection{Ablation Study}
Table~\ref{fig:multi_ablat} details different design choices regarding our SSL pretraining.
Ablations are done on ViT-S trained over the public Cholec80.
We report results on the validation set after linear evaluation.
In Tab.2a), we see that the method is robust to the number of prototypes, though over-clustering~\cite{caron2018deep} with 1k prototypes is optimal.
In Tab.2b) and Tab.2c), we explore the effect of random and focal masking.
We see that 50\% random masking (i.e. we keep 98 tokens out of 196 for the global view) and using 4 local views gives the best of performance.
In Tab.2d) we study the effect of data augmentation.
SSL augmentation pipelines have been developed on ImageNet-1k~\cite{chen2020simple}, hence, it is important to re-evaluate these choices for medical images.
Surprisingly, we see that augmentations primarily found to work well on ImageNet-1k are also effective on laparoscopic videos (e.g. color jiterring and horizontal flips).
In Tab.2e), we look at the effect of the training length when starting from scratch or from a good SSL pretrained checkpoint on ImageNet-1k.
We observe that excellent performance is achieved with only 10 epochs of finetuning on medical data when starting from a strong DINO checkpoint~\cite{caron2021emerging}. Tab.2g) shows that ImageNet-1k DINO is a solid starting point compared to other alternatives~\cite{he2022masked,zhou2021ibot,touvron2022deit,chen2021empirical}.
Finally, Tab.2f) confirms the necessity of regularizing with Sinkhorn-Knopp and me-max to avoid representation collapse by encouraging the use of all prototypes.

{\renewcommand{\arraystretch}{1.7}%
\begin{table*}[htbp]
\scriptsize
\centering
\caption{ Ablation study of different design choices (default setting is highlighted).}
\begin{tabularx}{0.97\textwidth}{ccccc|ccccc|ccc}
\hline
\multicolumn{5}{l|}{\cellcolor{gray!10}  a) Number of prototypes}                                                                                                            & \multicolumn{5}{l|}{\cellcolor{gray!10} d) Data augmentation}                                                                                                                                                                     & \multicolumn{3}{l}{\cellcolor{gray!10} f) Avoiding collapse.}                                       \\ \hline
$K$                   & $10^1$                 & $10^2$                       & \colorbox{Light}{$10^3$}                 & $10^4$                 & color jit                                         & flip (hor)                                        & blur                                              & \multicolumn{2}{c|}{val}                          & \colorbox{Light}{SK+me-max}     & SK       & $\emptyset$                  \\ \hline
val                   & 65.4                   & 67.8                         & \colorbox{Light}{69.8}                   & 69.1                   & \cellcolor{Light}\checkmark & \cellcolor{Light}\checkmark & \cellcolor{Light}\checkmark & \multicolumn{2}{c|}{\cellcolor{Light}{69.8}} & \colorbox{Light}{69.8}          & 67.7     & 34.0                         \\ \cline{1-5} \cline{11-13} 
\multicolumn{5}{l|}{\cellcolor{gray!10} b) Effect of random masking}                                                                                                        & \checkmark                         & \checkmark                         &                                                   & \multicolumn{2}{c|}{69.8}                         & \multicolumn{3}{c}{\cellcolor{gray!10} g) ImNet-1k initialization}                                   \\ \cline{1-5} \cline{11-13} 
\%                    & 0                      & \colorbox{Light}{50}   & 70                                             & 90                     & \checkmark                         &                                                   & \checkmark                         & \multicolumn{2}{c|}{68.6}                         & \multicolumn{2}{l}{weights. (ViT-B/16)}          & val                          \\ \cline{1-5} \cline{11-13} 
val                   & 69.1                   & \colorbox{Light}{69.8} & 68.4                                           & 68.2                   &                                                   & \checkmark                         & \checkmark                         & \multicolumn{2}{c|}{67.4}                         & \multicolumn{2}{c}{MAE~\cite{he2022masked}}                          & 53.5                         \\ \cline{1-10}
\multicolumn{5}{l|}{\cellcolor{gray!10} c) Local crops (focal masking)}                                                                                                     & \multicolumn{5}{l|}{\cellcolor{gray!10} e) Training length}                                                                                                                                                                      & \multicolumn{2}{c}{Supervised~\cite{touvron2022deit}}                   & 63.1                         \\ \cline{1-10}
\#                    & 0                      & 2                            & \colorbox{Light}4                      & 8                      & epochs                                            & 10                                                & 100                                               & \colorbox{Light}{200}         & 500         & \multicolumn{2}{c}{MoCo-v3~\cite{chen2021empirical}}                      & 63.3                         \\
                      &                        &                              & \colorbox{Light}                       &                        & scratch                                           & 33.8                                              & 63.3                                              & 65.5                                & 66.5        & \multicolumn{2}{c}{iBOT~\cite{zhou2021ibot}}                         & 65.7                         \\
\multirow{-2}{*}{val} & \multirow{-2}{*}{67.7} & \multirow{-2}{*}{69.1}       & \multirow{-2}{*}{\colorbox{Light}{69.8}} & \multirow{-2}{*}{68.1} & \colorbox{Light}{SSL init}                  & 68.2                                              & 69.3                                              & \colorbox{Light}{69.8}        & 68.4        & \multicolumn{2}{c}{\cellcolor{Light}{DINO~\cite{caron2021emerging}}} & \cellcolor{Light}{65.9}\\ \hline
\end{tabularx}
\label{fig:multi_ablat}
\end{table*}}

\section{Conclusion}

This study showcases the use of Masked Siamese Networks to learn informative representations from large, unlabeled endoscopic datasets. The learnt representations lead to state-of-the-art results in identifying surgical phases of laparoscopic procedures and in optical characterization of colorectal polyps. Moreover, this methodology displays strong generalization, achieving comparable performance with just 50\% of labeled data compared to standard supervised training on the complete labeled datasets. This dramatically reduces the need for annotated medical data, thereby facilitating the development of AI methods for healthcare.

\newpage

%
%
%
\bibliographystyle{splncs04}
\bibliography{references}

\begin{thebibliography}{10}
\providecommand{\url}[1]{\texttt{#1}}
\providecommand{\urlprefix}{URL }
\providecommand{\doi}[1]{https://doi.org/#1}

\bibitem{antonelli2023current}
Antonelli, G., Rizkala, T., Iacopini, F., Hassan, C.: Current and future
  implications of artificial intelligence in colonoscopy. Annals of
  Gastroenterology  \textbf{36}(2),  114--122 (2023)

\bibitem{assran2022masked}
Assran, M., Caron, M., Misra, I., Bojanowski, P., Bordes, F., Vincent, P.,
  Joulin, A., Rabbat, M., Ballas, N.: Masked siamese networks for
  label-efficient learning. In: ECCV (2022)

\bibitem{byrne2017will}
Byrne, M.F., Shahidi, N., Rex, D.K.: Will computer-aided detection and
  diagnosis revolutionize colonoscopy? Gastroenterology  \textbf{153}(6),
  1460--1464 (2017)

\bibitem{caron2018deep}
Caron, M., Bojanowski, P., Joulin, A., Douze, M.: Deep clustering for
  unsupervised learning of visual features. In: ECCV (2018)

\bibitem{caron2020unsupervised}
Caron, M., Misra, I., Mairal, J., Goyal, P., Bojanowski, P., Joulin, A.:
  Unsupervised learning of visual features by contrasting cluster assignments.
  NeurIPS  (2020)

\bibitem{caron2021emerging}
Caron, M., Touvron, H., Misra, I., J{\'e}gou, H., Mairal, J., Bojanowski, P.,
  Joulin, A.: Emerging properties in self-supervised vision transformers. In:
  ICCV (2021)

\bibitem{chen2020simple}
Chen, T., Kornblith, S., Norouzi, M., Hinton, G.: A simple framework for
  contrastive learning of visual representations. In: ICML (2020)

\bibitem{chen2021exploring}
Chen, X., He, K.: Exploring simple siamese representation learning. In: CVPR
  (2021)

\bibitem{chen2021empirical}
Chen, X., Xie, S., He, K.: An empirical study of training self-supervised
  vision transformers. In: ICCV (2021)

\bibitem{cohen2021has}
Cohen, R., Blau, Y., Freedman, D., Rivlin, E.: It has potential:
  Gradient-driven denoisers for convergent solutions to inverse problems.
  Advances in Neural Information Processing Systems  \textbf{34},  18152--18164
  (2021)

\bibitem{cohen2021regularization}
Cohen, R., Elad, M., Milanfar, P.: Regularization by denoising via fixed-point
  projection ({RED-PRO}). SIAM Journal on Imaging Sciences  \textbf{14}(3),
  1374--1406 (2021)

\bibitem{da2019self}
da~Costa~Rocha, C., Padoy, N., Rosa, B.: Self-supervised surgical tool
  segmentation using kinematic information. In: 2019 International Conference
  on Robotics and Automation (ICRA). pp. 8720--8726. IEEE (2019)

\bibitem{czempiel2020tecno}
Czempiel, T., Paschali, M., Keicher, M., Simson, W., Feussner, H., Kim, S.T.,
  Navab, N.: Tecno: Surgical phase recognition with multi-stage temporal
  convolutional networks. In: Medical Image Computing and Computer Assisted
  Intervention--MICCAI 2020: 23rd International Conference, Lima, Peru, October
  4--8, 2020, Proceedings, Part III 23. pp. 343--352. Springer (2020)

\bibitem{dayyeh2015asge}
Dayyeh, B.K.A., Thosani, N., Konda, V., Wallace, M.B., Rex, D.K., Chauhan,
  S.S., Hwang, J.H., Komanduri, S., Manfredi, M., Maple, J.T., et~al.: Asge
  technology committee systematic review and meta-analysis assessing the asge
  pivi thresholds for adopting real-time endoscopic assessment of the histology
  of diminutive colorectal polyps. Gastrointestinal endoscopy  \textbf{81}(3),
  502--e1 (2015)

\bibitem{dehghani2021scenic}
Dehghani, M., Gritsenko, A., Arnab, A., Minderer, M., Tay, Y.: Scenic: A jax
  library for computer vision research and beyond. In: CVPR. pp. 21393--21398
  (2022)

\bibitem{dosovitskiy2020image}
Dosovitskiy, A., Beyer, L., Kolesnikov, A., Weissenborn, D., Zhai, X.,
  Unterthiner, T., Dehghani, M., Minderer, M., Heigold, G., Gelly, S., et~al.:
  An image is worth 16x16 words: Transformers for image recognition at scale.
  arXiv preprint arXiv:2010.11929  (2020)

\bibitem{golany2022artificial}
Golany, T., Aides, A., Freedman, D., Rabani, N., Liu, Y., Rivlin, E., Corrado,
  G.S., Matias, Y., Khoury, W., Kashtan, H., et~al.: {AI} for phase recognition
  in complex laparoscopic cholecystectomy. Surgical Endoscopy pp.~1--9 (2022)

\bibitem{goldbraikh2023bounded}
Goldbraikh, A., Avisdris, N., Pugh, C.M., Laufer, S.: Bounded future {MS-TCN}++
  for surgical gesture recognition. In: ECCV 2022 Workshops, October 23--27,
  2022, Proceedings, Part III. pp. 406--421. Springer (2023)

\bibitem{hassan2021performance}
Hassan, C., Spadaccini, M., Iannone, A., Maselli, R., Jovani, M., Chandrasekar,
  V.T., Antonelli, G., Yu, H., Areia, M., Dinis-Ribeiro, M., et~al.:
  Performance of artificial intelligence in colonoscopy for adenoma and polyp
  detection: a systematic review and meta-analysis. Gastrointestinal endoscopy
  \textbf{93}(1),  77--85 (2021)

\bibitem{he2022masked}
He, K., Chen, X., Xie, S., Li, Y., Doll{\'a}r, P., Girshick, R.: Masked
  autoencoders are scalable vision learners. In: CVPR (2022)

\bibitem{intrator2023reid}
Intrator, Y., Aizenberg, N., Livne, A., Rivlin, E., Goldenberg, R.:
  Self-supervised polyp re-identification in colonoscopy. arXiv preprint
  arXiv:2306.08591  (2023)

\bibitem{joulin2012convex}
Joulin, A., Bach, F.: A convex relaxation for weakly supervised classifiers.
  arXiv preprint arXiv:1206.6413  (2012)

\bibitem{katzir2022estimating}
Katzir, L., Veikherman, D., Dashinsky, V., Goldenberg, R., Shimshoni, I.,
  Rabani, N., Cohen, R., Kelner, O., Rivlin, E., Freedman, D.: Estimating
  withdrawal time in colonoscopies. In: ECCV. pp. 495--512. Springer (2022)

\bibitem{kutiel2023conformal}
Kutiel, G., Cohen, R., Elad, M., Freedman, D., Rivlin, E.: Conformal prediction
  masks: Visualizing uncertainty in medical imaging. In: ICLR 2023 Workshop on
  Trustworthy Machine Learning for Healthcare (2023)

\bibitem{livovsky2021detection}
Livovsky, D.M., Veikherman, D., Golany, T., Aides, A., Dashinsky, V., Rabani,
  N., Shimol, D.B., Blau, Y., Katzir, L., Shimshoni, I., Liu, Y., Segol, O.,
  Goldin, E., Corrado, G., Lachter, J., Matias, Y., Rivlin, E., Freedman, D.:
  Detection of elusive polyps using a large-scale artificial intelligence
  system (with videos). Gastrointestinal Endoscopy  \textbf{94}(6),  1099--1109
  (2021)

\bibitem{ou2021polyp}
Ou, S., Gao, Y., Zhang, Z., Shi, C.: {Polyp-YOLOv5-Tiny}: A lightweight model
  for real-time polyp detection. In: International Conference on Information
  Technology, Big Data and Artificial Intelligence (ICIBA). vol.~2, pp.
  1106--1111 (2021)

\bibitem{ramesh2023dissecting}
Ramesh, S., Srivastav, V., Alapatt, D., Yu, T., Murali, A., Sestini, L., Nwoye,
  C.I., Hamoud, I., Sharma, S., Fleurentin, A., et~al.: Dissecting
  self-supervised learning methods for surgical computer vision. Medical Image
  Analysis  \textbf{88},  102844 (2023)

\bibitem{ross2018exploiting}
Ross, T., Zimmerer, D., Vemuri, A., Isensee, F., Wiesenfarth, M., Bodenstedt,
  S., Both, F., Kessler, P., Wagner, M., M{\"u}ller, B., et~al.: Exploiting the
  potential of unlabeled endoscopic video data with self-supervised learning.
  International journal of computer assisted radiology and surgery
  \textbf{13},  925--933 (2018)

\bibitem{russakovsky2015imagenet}
Russakovsky, O., Deng, J., Su, H., Krause, J., Satheesh, S., Ma, S., Huang, Z.,
  Karpathy, A., Khosla, A., Bernstein, M., et~al.: Imagenet large scale visual
  recognition challenge. IJCV  (2015)

\bibitem{sestini2021kinematic}
Sestini, L., Rosa, B., De~Momi, E., Ferrigno, G., Padoy, N.: A kinematic
  bottleneck approach for pose regression of flexible surgical instruments
  directly from images. IEEE Robotics and Automation Letters  \textbf{6}(2),
  2938--2945 (2021)

\bibitem{touvron2022deit}
Touvron, H., Cord, M., J{\'e}gou, H.: {DeIT III}: Revenge of the {ViT}. arXiv
  preprint arXiv:2204.07118  (2022)

\bibitem{twinanda2016endonet}
Twinanda, A.P., Shehata, S., Mutter, D., Marescaux, J., De~Mathelin, M., Padoy,
  N.: Endonet: a deep architecture for recognition tasks on laparoscopic
  videos. IEEE transactions on medical imaging  \textbf{36}(1),  86--97 (2016)

\bibitem{polypset2021}
Wang, G.: Replication data for: Colonoscopy polyp detection and classification:
  Dataset creation and comparative evaluations. Harvard Dataverse  (2021),
  \url{https://doi.org/10.7910/DVN/FCBUOR}

\bibitem{zhou2021ibot}
Zhou, J., Wei, C., Wang, H., Shen, W., Xie, C., Yuille, A., Kong, T.: ibot:
  Image bert pre-training with online tokenizer. arXiv preprint
  arXiv:2111.07832  (2021)

\end{thebibliography}
%





\newpage

\section*{Supplementary materials}





\section{Laparoscopic Datasets}
\begin{minipage}{0.48\textwidth}
\includegraphics[width=\textwidth]{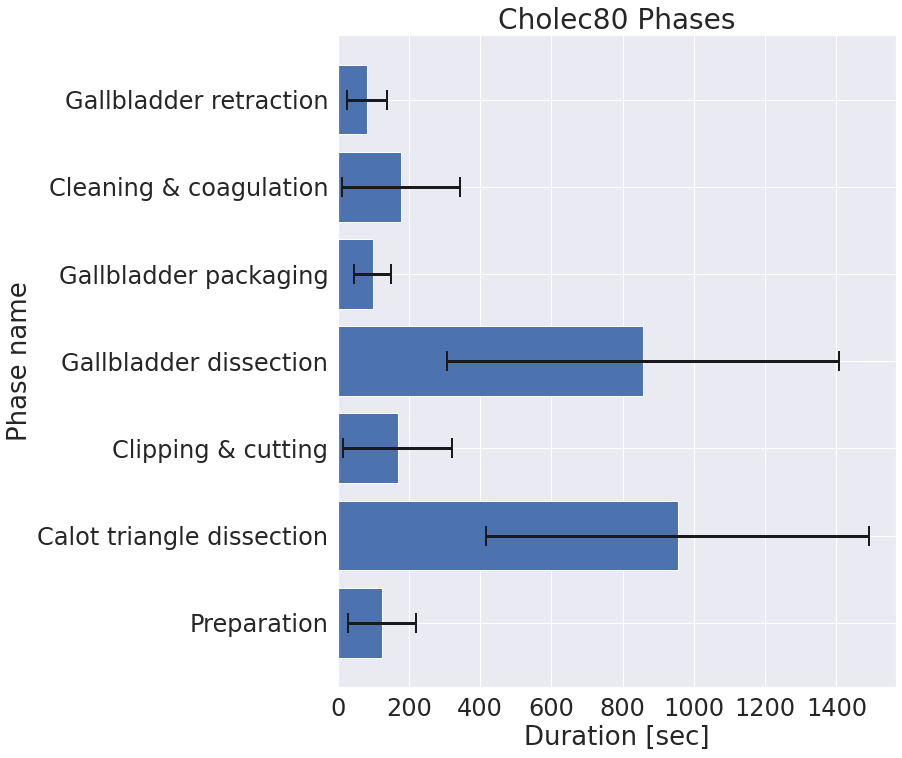}
\captionof{figure}{The phases of the \textit{public} laparoscopic dataset Cholec80~\cite{twinanda2016endonet}. We report the mean and standard deviation of each phase's duration in seconds.}
\end{minipage}
\hfill
\begin{minipage}{0.48\textwidth}
\includegraphics[width=\textwidth]{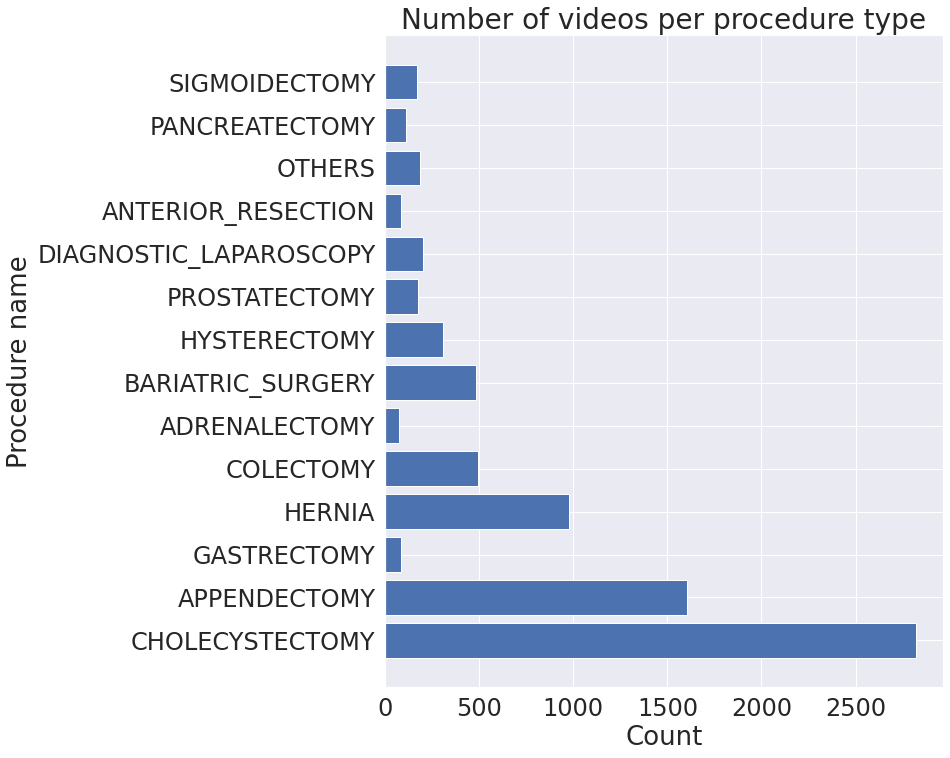}
\captionof{figure}{A description of the procedures that comprise the \textit{private} laparoscopic dataset. We report the number of videos per procedure type.}
\end{minipage}

\section{Implementation details}
\subsection{Masked Siamese Networks}
\vspace{-1cm}
\begin{table}[h!]
    \caption{Description of the different ViT~\cite{dosovitskiy2020image} variants used in our experiments.}
    \vspace{3mm}
    \centering
    \begin{tabular}{c|c|c|c|c|c}
    \toprule
        Model & Layers & Hidden dim size & MLP size & Num heads & Num params \\ \midrule
        ViT-S & 12 & 384 & 1536 & 6 & 23M \\ 
        ViT-B & 12 & 768 & 3072 & 12 & 86M \\ 
        ViT-L & 24 & 1024 & 4096 & 16 & 307M \\ 
    \bottomrule
    \end{tabular}
    \label{tab:vit}
\end{table}
\vspace{-1.5cm}
\begin{table}[h!]
    \caption{Hyper-parameters used for MSN training. For data augmentation we apply random resized crop, horizontal flipping and color jittering (following~\cite{chen2020simple}).The input pipeline is similar to that of MSN original paper~\cite{assran2022masked}. We report results for the best performing models across the validation set.}
    \vspace{3mm}
    \centering
    \begin{tabular}{ccccc}
    \toprule
        Learning rate & Weight decay & Seq. length & Batch size & Num epochs \\ \midrule
        \{3e-1, 1e-3, 3e-4, 1e-4\} & 0.01 & \{40, 96, 196\} & 1024 & 200 \\
    \bottomrule
    \end{tabular}
    \vspace{3mm}
    \label{tab:msn}
\end{table}


\newpage
\subsection{Downstream experiments}
\vspace{-1cm}

\begin{table}[htbp]
    \caption{Hyper-parameters used for Cholec80 per-frame recognition task. For all downstream tasks we report results for the best performing models across the validation sets.}
    \vspace{3mm}
    \centering
    \begin{tabular}{p{1in}p{1in}p{1in}cc}
    \toprule
    Optimizer & Learning rate & Weight decay & Batch size & Num epochs  \\ \midrule
    \{Adam, AdamW, SGD\} & \{5e-3, 5e-3, 1e-4, 5e-4, 1e-5\} & \{0, 1e-5, 5e-5, 1e-4\} & 265 & 30\\
    \bottomrule
    \end{tabular}
    \label{tab:cholec_frame_ft}
    \caption{Hyper-parameters used for Cholec80 temporal recognition task.}
    \vspace{3mm}
    \begin{tabular}{ccccc}
    \toprule
    Optimizer & Learning rate & Weight decay & Batch size & Num epochs  \\ \midrule
    \{Adam, AdamW, SGD\} & \{1e-3, 5e-4, 3e-5, 1e-4\} & \{0, 1e-5, 5e-5\} & 8 & 30\\
    \bottomrule
    \end{tabular}
    \label{tab:cholec_temporal_ft}
    \caption{Hyper-parameters used for PolypSet polyp characterization task.}
    \vspace{3mm}
    \begin{tabular}{ccccc}
    \toprule
    Optimizer & Learning rate & Weight decay & Batch size & Num epochs  \\ \midrule
    \{Adam, AdamW, SGD\} & \{1e-3, 5e-4, 3e-5, 1e-4\} & \{0, 1e-5, 5e-5\} & 256 & 50\\
    \bottomrule
    \end{tabular}
    \label{tab:polypset_frame_le}

\end{table}




\end{document}